\documentclass[11pt,a4paper]{article}
\usepackage[margin=26mm]{geometry}
\usepackage[T1]{fontenc}
\usepackage[utf8]{inputenc}
\usepackage{lmodern}
\usepackage{microtype}
\usepackage{amsmath,amssymb,amsfonts}
\usepackage{algorithm,algorithmic}
\usepackage{graphicx,booktabs,tabularx,caption}
\usepackage[section]{placeins} % keep figures in their section
\usepackage[numbers,sort&compress]{natbib}
\usepackage{authblk}
\usepackage{url}
\usepackage{xcolor}
\usepackage[hidelinks]{hyperref}
\title{CARE: Contrastive Representation Learning with Environment-Invariance Regularization for Behavioural Intention Prediction}
\author[1]{Shang Gao\thanks{ORCID: \href{https://orcid.org/0009-0005-5474-1888}{0009-0005-5474-1888}}}
\author[1]{Zhixin Huang\thanks{ORCID: \href{https://orcid.org/0000-0002-6124-0206}{0000-0002-6124-0206}}}
\author[1]{Franz G\"otz-Hahn\thanks{ORCID: \href{https://orcid.org/0000-0003-3465-5040}{0000-0003-3465-5040}}}
\author[2]{Konrad Doll\thanks{ORCID: \href{https://orcid.org/0000-0002-3746-2319}{0000-0002-3746-2319}}}
\author[1]{Bernhard Sick\thanks{ORCID: \href{https://orcid.org/0000-0001-9467-656X}{0000-0001-9467-656X}}}
\affil[1]{Intelligent Embedded Systems, University of Kassel, Germany\\\texttt{\{sgao,
zhixin.huang,bsick,franz.goetz-hahn\}@uni-kassel.de}}
\affil[2]{Aschaffenburg University of Applied Sciences, Germany\\\texttt{konrad.doll@th-ab.de}}
\date{}
\hypersetup{
  pdftitle={CARE: Contrastive Representation Learning with Environment-Invariance Regularization for Behavioural Intention Prediction},
  pdfauthor={Shang Gao, Zhixin Huang, Franz Goetz-Hahn, Konrad Doll, Bernhard Sick}
}
\begin{document}
\maketitle

\begin{abstract}
Trajectory-based behavioural intention prediction faces environmental variation, and representations learned with a pooled contrastive objective may retain environment-specific regularities. We present CARE, a self-supervised trajectory-learning framework that couples temporally asymmetric contrastive views with an invariant-risk-minimisation (IRM)-style gradient penalty on environment-wise contrastive risks. Each input window yields an anchor, a partially masked positive view, and a temporally perturbed negative view. Masked reconstruction and observability-aware evidence consistency complement this objective by promoting temporal information retention under partial input. On an IMPTC-derived four-class intention task, CARE achieves 0.840 accuracy and 0.023 expected calibration error (ECE) under IID fine-tuning, and 0.806 accuracy with a Brier score of 0.308 on the reported held-out environment. Under frozen-encoder linear probing, reconstruction provides the largest individual held-out accuracy gain, whereas the complete objective yields the lowest held-out ECE (0.049) among tested variants. These results support the combined objective as a competitive representation-learning approach under the evaluated shift, while the IRM-style regulariser is not taken as evidence of formal invariance.
\end{abstract}


\section{Introduction}
\label{sec:intro}
Behavioural Intention Prediction (BIP) estimates a road user's upcoming manoeuvre from a short observed motion history. For vulnerable road users (VRUs), including pedestrians and cyclists, such estimates are relevant to proactive planning in autonomous driving~\cite{Fang2024bipsurvey}. Models are expected not only to distinguish intentions such as moving straight, turning or stopping, but also to retain predictive value when conditions at deployment differ from those seen during training.

Trajectory data are often heterogeneous: lighting, precipitation and time of day can vary across recordings. Standard supervised learning and conventional self-supervised pretraining do not, by themselves, require an encoder to avoid environment-specific shortcuts. This motivates a testable methodological question: \emph{can contrastive trajectory learning be regularized toward shared behaviour across predefined environments without sacrificing intention-prediction performance?} Invariance here is a learning objective, not a property established by the use of an IRM penalty alone~\cite{arjovsky2020invariantriskminimization}.

CARE addresses this question by combining temporally asymmetric trajectory views with an environment-wise penalty on contrastive risk. For each input window, an anchor is compared with a masked positive and a temporally corrupted negative. The earlier part of the \emph{observed input window} remains fixed while the later part is perturbed. The distinction is important: the perturbed segment is still observed input and is not the separate future interval used to derive the intention label. Contrastive learning encourages discrimination between these views; an IRM-inspired objective penalizes disparities in optimality of a shared projection head across training environments.

To complement the environment-regularised contrastive objective, CARE also reconstructs masked motion steps and constrains the evidential head across differently observed views. The resulting joint objective couples discriminative trajectory structure, environment-wise regularisation, and observability-dependent auxiliary learning without assuming that a single constraint can guarantee robustness.

The contributions of this study are as follows:
\begin{enumerate}
\item \textbf{Environment-regularized contrastive training.} We formulate a contrastive loss over asymmetric motion views and add an IRM-style gradient penalty on contrastive risks computed within precipitation/time-of-day environment groups.
\item \textbf{A controlled BIP evaluation.} We retain the original IMPTC-derived four-class proxy-label task and report IID and selected held-out-environment results for the same baseline and CARE configurations.
\item \textbf{A component-level reliability assessment.} Using full-model evaluations and frozen-encoder ablations, we distinguish the accuracy contribution of reconstruction from the probability-quality improvement associated with the combined objective, without attributing unverified gains to IRM alone.
\end{enumerate}

\begin{figure}[!tbp]
  \centering
  \includegraphics[width=0.99\linewidth]{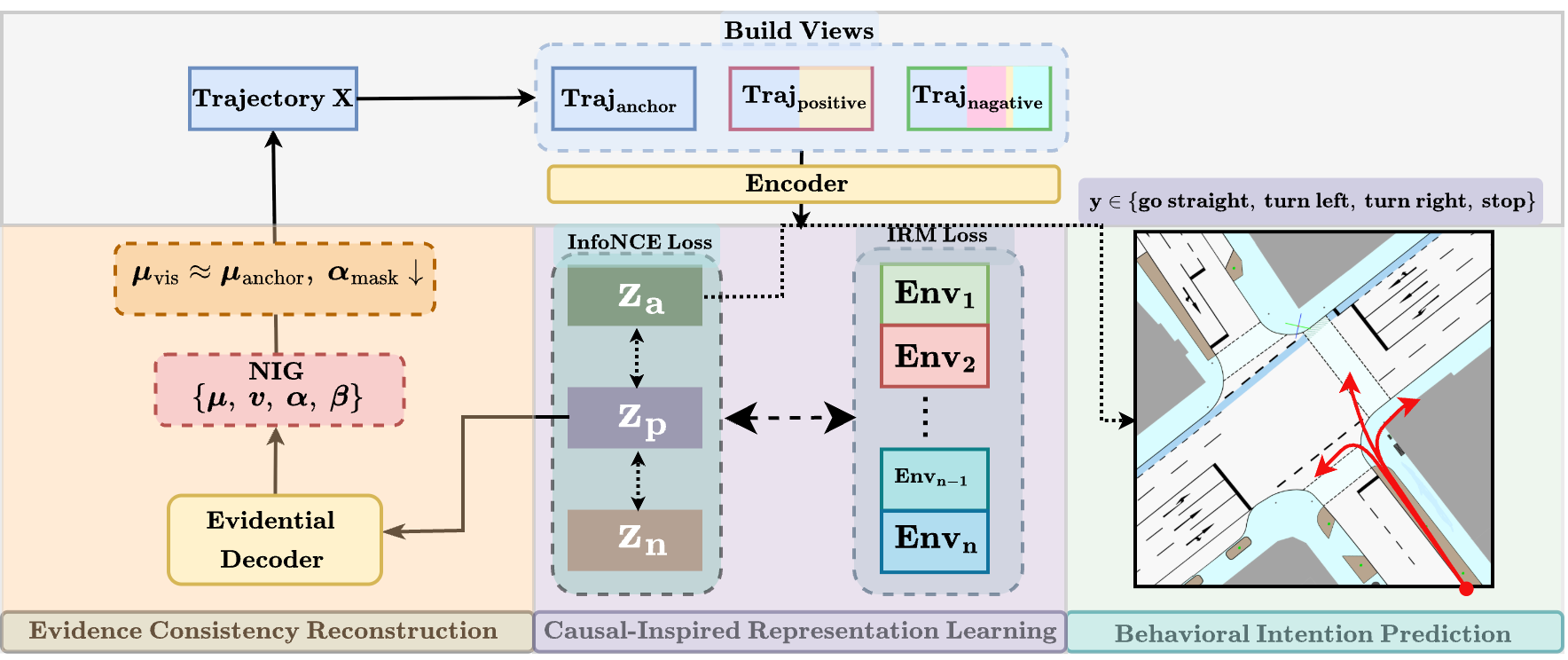}
  \caption{CARE combines temporally asymmetric contrastive views with an environment-wise IRM-style penalty; reconstruction and evidence consistency provide auxiliary regularization. The architecture and computation graph are retained from the original manuscript.}
  \label{fig:care}
\end{figure} 

\section{Related Work}

\subsection{Behavioural Intention Prediction and Trajectory Representations}
Behavioural intention prediction uses motion histories and contextual cues to distinguish the likely actions of road users~\cite{Fang2024bipsurvey,Zernetsch2024}. Learning from unlabelled motion is attractive when manually annotated trajectories are scarce. Contrastive predictive coding and TS2Vec demonstrate how self-supervised objectives can extract structure from sequential data~\cite{oord2019representationlearningcontrastivepredictive,yue2022ts2vecuniversalrepresentationtime}; masked autoencoding supplies a complementary reconstruction signal~\cite{li2023timaeselfsupervisedmaskedtime}. CARE uses both mechanisms but directs its main regularization toward the differences between training environments.

\subsection{Environment Shift and Invariance Regularization}
Domain generalization methods seek predictors that continue to work under distribution shifts. Domain-adversarial training suppresses domain-discriminative features~\cite{ganin2016domainadversarialtrainingneuralnetworks}, whereas invariant risk minimization (IRM) seeks representations for which one predictor is simultaneously adequate across environments~\cite{arjovsky2020invariantriskminimization}. Causal representation learning motivates, but does not guarantee, invariance under changes in nuisance conditions~\cite{schölkopf2021causalrepresentationlearning}. CARE adapts an IRM-style gradient penalty to an environment-wise contrastive risk instead of applying it directly to supervised intention classification. It is therefore an \emph{invariance-regularized objective}, not a proof that the learned embeddings contain only causal or invariant factors.

\subsection{Reconstruction and Observability-Aware Regularisation}
Masked reconstruction complements contrastive learning by encouraging an encoder to retain local trajectory information even when parts of its input are missing~\cite{li2023timaeselfsupervisedmaskedtime}. Evidential learning introduces structured uncertainty parameters~\cite{sensoy2018evidentialdeeplearningquantify,amini2020deepevidentialregression}; their interpretation under partial observability motivates regularisation of the response to missing context. In particular, the visible-mean consistency and masked-evidence principles developed for evidential masked pretraining in eCoMAE~\cite{Gao2026ecoMAE} motivate CARE's auxiliary constraints. Work on observability-dependent uncertainty, including MORSE~\cite{Gao2026MORSE}, further illustrates the importance of relating uncertainty claims to measurable changes in input context. CARE incorporates observability-sensitive constraints into a temporally asymmetric, environment-regularised contrastive objective and evaluates prediction and classification calibration with accuracy, expected calibration error (ECE), negative log-likelihood (NLL), and Brier score~\cite{guo2017calibrationmodernneuralnetworks}.

\section{Method}
\label{sec:method}
CARE targets environment-regularized contrastive representation learning for BIP. Table~\ref{tab:notation} summarizes notation. An input window $\mathbf{X}\in\mathbb{R}^{L\times F}$ and an environment label $e\in\{1,\ldots,E\}$ are available at pretraining time. The learned encoder $f_\theta$ is subsequently evaluated on four rule-derived intention classes: \textit{straight}, \textit{left}, \textit{right} and \textit{stop}. No causal graph or physical causal effect is estimated.

Three components appear in the objective: temporally asymmetric contrastive views, an environment-wise IRM-style contrastive penalty, and auxiliary masked reconstruction/evidence consistency. The components and their loss equations are defined below. Here, $L_p$ and $L_f$ refer to the early and late \emph{input} segments; the later input segment must not be confused with the separate prediction horizon $H$ used to construct proxy intention labels.

\begin{table}[!tbp]
\centering
\caption{Notation.}
\label{tab:notation}
\small
\setlength{\tabcolsep}{4pt}
\renewcommand{\arraystretch}{1.08}
\begin{tabularx}{\columnwidth}{@{}lX@{}}
\toprule
Symbol & Meaning \\
\midrule
$\mathbf{X}\in\mathbb{R}^{L\times F}$ & input trajectory window, length $L$, feature dim $F$ \\
$\mathbf{x}_t\in\mathbb{R}^{F}$ & feature vector at time step $t$ \\
$L_p$ & past length, $L_f=L-L_p$ future length \\
$\mathbf{A},\mathbf{P},\mathbf{N}$ & anchor, masked positive, corrupted negative views \\
$\mathcal{M}\subset\{1,\ldots,L\}$ & masked time index set used to form $\mathbf{P}$ \\
$\rho,\; b$ & mask rate, block length \\
$f_\theta$ & encoder, $\mathbf{Z}=f_\theta(\mathbf{X})\in\mathbb{R}^{L\times D}$ \\
$h_w$ & projection head for contrastive vectors $\mathbf{z}$ \\
$e\in\{1,\ldots,E\}$ & environment index, $E$ number of environments \\
$\mathcal{B},\;\mathcal{B}_e$ & training batch and its subset in environment $e$ \\
$d_\phi$ & decoder for reconstruction $\hat{\mathbf{X}}=d_\phi(\mathbf{Z}_p)$ \\
$u_\psi$ & evidential head outputting NIG parameters $\boldsymbol{\mu},\mathbf{v},\boldsymbol{\alpha},\boldsymbol{\beta}$ \\
$\tau,\;\lambda_{\cdot}$ & InfoNCE temperature and loss weights \\
\bottomrule
\end{tabularx}
\end{table}

\subsection{Temporally Asymmetric Contrastive Views}
\label{subsec:views}
CARE forms anchor, positive and negative views $\mathbf{A},\mathbf{P},\mathbf{N}\in\mathbb{R}^{L\times F}$ from the same input window. The anchor is unmodified. Positive views mask contiguous blocks within the later input segment $\{L_p+1,\ldots,L\}$, replacing the selected entries with a learnable token. Negative views apply one of three temporal corruptions to this same segment: rotation, reversal or swapping across a batch. The early $L_p$ input steps remain identical across all three views.

The mask ratio $\rho$ and block length $b$ determine the positive-view perturbation. The purpose of this asymmetry is to preserve a common earlier context while making the later input portion informative for distinguishing views; it is not an intervention on a physical cause, nor evidence that the downstream model is protected from future-label leakage. Algorithm~\ref{alg:views} specifies the view construction procedure.

\begin{algorithm}[t]
\caption{Temporally asymmetric contrastive views (original CARE procedure)}
\label{alg:views}
\small
\begin{algorithmic}[1]
\STATE \textbf{Input:} $\mathbf{X}\in\mathbb{R}^{L\times F}$, past length $L_p$, mask rate $\rho$, block length $b$, mask token $\mathbf{m}$
\STATE \textbf{Anchor:} $\mathbf{A}\leftarrow \mathbf{X}$
\STATE \textbf{Positive:} $\mathbf{P}\leftarrow \mathbf{X}$, $M\leftarrow \emptyset$
\STATE Partition later-input indices $\{L_p{+}1,\dots,L\}$ into consecutive blocks of length $b$
\STATE Uniformly sample $\lceil \rho \cdot \frac{L-L_p}{b} \rceil$ blocks without replacement
\FOR{each sampled block with indices $\{s,\dots,e\}$}
  \STATE $M\leftarrow M\cup\{s,\dots,e\}$; set $\mathbf{P}_{t}\leftarrow \mathbf{m}$ for all $t\in\{s,\dots,e\}$
\ENDFOR
\STATE \textbf{Negative:} $\mathbf{N}\leftarrow \mathbf{X}$
\STATE apply one random corruption to the late segment $\mathbf{N}_{L_p+1:L}$:
\STATE \hspace{0.8em} \textit{Rotate} (temporal roll), \textit{Swap} (batch roll), or \textit{Reverse} (time flip)
\STATE \textbf{Output:} $\mathbf{A},\mathbf{P},\mathbf{N},M$
\end{algorithmic}
\end{algorithm}

\subsection{Environment-Invariance Regularization of Contrastive Risk}
\label{subsec:ci}
The encoder processes the three views, producing latent sequences
$\mathbf{Z}_a=f_\theta(\mathbf{A})$, $\mathbf{Z}_p=f_\theta(\mathbf{P})$, $\mathbf{Z}_n=f_\theta(\mathbf{N})\in\mathbb{R}^{L\times D}$.
A projection head produces contrastive vectors
$\mathbf{z}_a,\mathbf{z}_p,\mathbf{z}_n\in\mathbb{R}^{D}$.
The InfoNCE loss aligns anchor and positive while repelling the negative:
\begin{equation}
L_{\text{nce}}
= -\log
\frac{\exp(\mathrm{sim}(\mathbf{z}_a,\mathbf{z}_p)/\tau)}
{\exp(\mathrm{sim}(\mathbf{z}_a,\mathbf{z}_p)/\tau)+\exp(\mathrm{sim}(\mathbf{z}_a,\mathbf{z}_n)/\tau)} ,
\end{equation}
where $\mathrm{sim}$ is cosine similarity and $\tau$ is the temperature.

To discourage reliance on environment-specific shortcuts, CARE applies an IRM-style penalty to the environment-wise contrastive risk.
Each sample is associated with an environment index $e\in\{1,\dots,E\}$.
Let $\mathcal{D}_e$ denote the set of all training samples from environment $e$.
During training, a batch of samples $\mathcal{B}\subset\cup_{e=1}^E\mathcal{D}_e$ is drawn and partitioned by environment.
Let $\mathcal{B}_e=\{\mathbf{x}\in\mathcal{B}:\mathrm{env}(\mathbf{x})=e\}$ denote the subset of the current batch from environment $e$.
The environment-wise contrastive risk is estimated as
\begin{equation}
\widehat{R}_e(\theta,w)=\frac{1}{|\mathcal{B}_e|}\sum_{\mathbf{x}\in\mathcal{B}_e}\ell_{\text{nce}}(\theta,w;\mathbf{x}).
\end{equation}

CARE adopts an IRM-style gradient penalty intended to favour a shared projection mechanism across training environments; minimizing this surrogate does not itself establish causal or statistical invariance.
Following the IRM principle, the penalty is measured by the gradient of each environment risk with respect to the head parameters,
evaluated at a fixed reference head $w_0$,
\begin{equation}
L_{\text{irm}}
=\sum_{e=1}^{E}\left\|\nabla_w \widehat{R}_e(\theta,w)\right\|_2^2\Big|_{w=w_0}.
\label{eq:irm}
\end{equation}
Here $w_0$ is a shared and frozen initialization, and is not updated during training.
Minimising $L_{\text{irm}}$ updates the encoder parameters $\theta$ such that the same simple head can achieve low contrastive risk in all environments.
The penalty encourages shared-head compatibility across the observed training groups, although stability on unobserved groups requires empirical assessment.

The environment-regularized contrastive objective is
\begin{equation}
L_{\text{ci}}=\lambda_{\text{nce}}L_{\text{nce}}+\lambda_{\text{irm}}L_{\text{irm}},
\end{equation}
where $L_{\text{nce}}$ is averaged over samples in the batch $\mathcal{B}$. The term $L_{\text{irm}}$ penalises environment-wise gradients as a surrogate for a common predictor; it does not enforce exact invariance across either the observed groups or unseen environments.

\subsection{Auxiliary Reconstruction and Evidence Consistency}
\label{evcr}
In addition to contrastive learning, CARE retains masked reconstruction and an observability-dependent evidential consistency penalty on the positive view $\mathbf{P}$.
A decoder $d_{\varphi}$ reconstructs the input sequence from the latent representation $\mathbf{Z}_p=f_{\theta}(\mathbf{P})$,
\begin{equation}
\hat{\mathbf{X}} = d_{\varphi}(\mathbf{Z}_p), \qquad \hat{\mathbf{x}}_t \in \mathbb{R}^{F}.
\end{equation}
Let $\mathcal{M}\subset\{1,\dots,L\}$ denote the set of masked time indices used to form $\mathbf{P}$.
The reconstruction loss is computed only on masked steps and encourages the encoder to preserve fine-grained temporal structure under partial observability,
\begin{equation}
L_{\text{rec}}
= \frac{1}{|\mathcal{M}|}
\sum_{t\in\mathcal{M}}
\bigl\|
\mathbf{x}_t-\hat{\mathbf{x}}_t
\bigr\|_2^2 .
\end{equation}

Reconstruction alone does not constrain the outputs of an evidential head.
As an auxiliary regularizer, CARE also uses a head $u_{\psi}$ that outputs Normal-Inverse-Gamma parameters for each time step,
\begin{equation}
(\boldsymbol{\mu}, \mathbf{v}, \boldsymbol{\alpha}, \boldsymbol{\beta})
= u_{\psi}(\mathbf{Z}), \qquad
\boldsymbol{\mu}_t\in\mathbb{R}^{F},\ \boldsymbol{\alpha}_t\in\mathbb{R}^{F}.
\end{equation}
The parameter $\boldsymbol{\mu}_t$ represents a predictive mean, while $\boldsymbol{\alpha}_t$ is the NIG shape parameter regularised at masked positions. These observability-dependent constraints follow the motivation of evidential masked pretraining~\cite{Gao2026ecoMAE}. Because the specified auxiliary objective does not include an NIG likelihood, the parameters are used as training signals rather than interpreted as validated calibrated uncertainty estimates.

The evidence-consistency regularizer relates outputs to observability through two complementary constraints defined on the anchor view $\mathbf{A}$ and the positive view $\mathbf{P}$.
Let $(\boldsymbol{\mu}^{a},\boldsymbol{\alpha}^{a})=u_{\psi}(\mathbf{Z}_a)$ and $(\boldsymbol{\mu}^{p},\boldsymbol{\alpha}^{p})=u_{\psi}(\mathbf{Z}_p)$, where $\mathbf{Z}_a=f_{\theta}(\mathbf{A})$.
Let $\bar{\mathcal{M}}=\{1,\dots,L\}\setminus\mathcal{M}$ denote visible indices.
First, on visible steps, predictive means are encouraged to agree between the anchor and the masked view,
\begin{equation}
\mathcal{L}_{\mu}^{\text{vis}}
=
\frac{1}{|\bar{\mathcal{M}}|}
\sum_{t\in\bar{\mathcal{M}}}
\bigl\|
\boldsymbol{\mu}^{p}_t-\boldsymbol{\mu}^{a}_t
\bigr\|_2^2 .
\end{equation}
Second, on masked steps, large values of the NIG shape parameter are penalised,
\begin{equation}
\mathcal{L}_{\alpha}^{\text{mask}}
=
\frac{1}{|\mathcal{M}|}
\sum_{t\in\mathcal{M}}
\bigl\|
\boldsymbol{\alpha}^{p}_t
\bigr\|_1 .
\end{equation}
The evidence-consistency regulariser is then
\begin{equation}
\mathcal{L}_{\text{ecr}}=
\mathcal{L}_{\mu}^{\text{vis}}+\mathcal{L}_{\alpha}^{\text{mask}} .
\end{equation}

\subsection{Combined Training Objective}
CARE combines reconstruction and evidence consistency as an auxiliary objective:
\begin{equation}
L_{\text{cal}}
= \lambda_{\text{rec}} L_{\text{rec}}
+ \lambda_{\text{ecr}} L_{\text{ecr}}.
\end{equation}
Together with the environment-regularized contrastive objective $L_{\text{ci}}$ in Sec.~\ref{subsec:ci}, the total self-supervised objective is
\begin{equation}
L_{\text{ssl}}
= \sum_{k \in \mathcal{K}} \lambda_k L_k ,
\end{equation}
where $\mathcal{K} = \{\text{ci}, \text{cal}\}$ and $\lambda_k \ge 0$ are scalar weights.

CARE is trained under two alternative protocols: two-stage training and joint training.
In the two-stage protocol, the encoder $f_\theta$ is first pretrained by minimising $L_{\text{ssl}}$, and then frozen while a linear classifier is trained for intention prediction.
In the joint protocol, the same $L_{\text{ssl}}$ is added as a regulariser to the supervised cross-entropy loss, enabling end-to-end optimisation while preserving the same objective components.

\section{Experiments: Predictive Performance and Component Trade-offs}
\label{sec:experiments}

\subsection{Dataset Construction and Task}
\label{sec:dataset}

A window level behavioural intention prediction dataset is constructed from IMPTC trajectory data.
The raw data are frame wise tables grouped by an entity identifier, together with an entity level metadata table.
For each identifier, the sequence is sorted by a time field and samples are generated via sliding windows.
Given a sequence of length $L$, a start index $s$ is advanced with stride $S$, a history window $[s, s+L)$ is extracted as the model input,
and the subsequent segment $[s+L, s+L+H)$ is used to derive manoeuvre cues for label generation.
Unless otherwise stated, the window length is $L=100$, the stride is $S=80$, and the future horizon for cue extraction is $H=50$.

The downstream task is a four class intention classification problem with labels:
$0$ straight, $1$ left turn, $2$ right turn, $3$ stop.

\paragraph{Input features}
Each sample contains a trajectory tensor $X \in \mathbb{R}^{L \times 7}$ with seven per timestep features in a fixed order:
$x$, $y$, speed magnitude, $z$ height, ground\_type, z\_offset, source\_type.
In addition, each sequence is associated with two context variables used for environment construction:
precipitation intensity and part of day.

\paragraph{Rule derived intention labels}
Window level intention annotations $y$ are derived from kinematic rules.
Stop is assigned if the mean speed over the last $K$ steps of the history window is below $v_{\text{stop}}$.
The configuration uses $v_{\text{stop}}=0.25$ and $K \in [3,10]$.
Otherwise, the signed angle $\Delta\theta$ between the last displacement in the history and the displacement from the history end to the future end is computed.
Left is assigned if $\Delta\theta > \theta$, right is assigned if $\Delta\theta < -\theta$, and straight is assigned otherwise.
The configuration uses $\theta = 12.5^{\circ}$.
These are proxy labels, and their utility is assessed through calibration and robustness metrics in addition to accuracy.

\paragraph{Dataset scale and label distribution}
Tables~\ref{tab:dataset_overview} and~\ref{tab:label_distribution} report dataset scale and label distribution.
All supervised models are trained with class-weighted cross-entropy to reduce bias towards frequent classes.
For evaluation, accuracy is used to enable a fair comparison with standard time-series baselines under identical settings.
Although accuracy can be influenced by imbalance, the comparisons remain consistent because the same metric and splits are used for all methods.

\begin{table}[t]
\centering
\caption{IMPTC BIP dataset overview and window construction settings.}
\label{tab:dataset_overview}
\small
\setlength{\tabcolsep}{4pt}
\renewcommand{\arraystretch}{1.11}
\begin{tabular*}{0.94\linewidth}{@{\extracolsep{\fill}}lcccccc@{}}
\toprule
Split & Samples & \shortstack{Unique\\ IDs} & \shortstack{Window length\\ $L$} & \shortstack{Past length\\ $L_p$} & \shortstack{Future length\\ $L_f$} & \shortstack{Stride\\ $S$} \\
\midrule
Train & 28272 & 3328 & 100 & 60 & 40 & 80 \\
Test  & 8155  & 1065 & 100 & 60 & 40 & 80 \\
\bottomrule
\end{tabular*}
\end{table}

\begin{table}[t]
\centering
\caption{Intention label distribution. Percentages are computed within each split.}
\label{tab:label_distribution}
\small
\setlength{\tabcolsep}{4pt}
\renewcommand{\arraystretch}{1.11}
\begin{tabular*}{0.94\linewidth}{@{\extracolsep{\fill}}lcccc@{}}
\toprule
Split & Straight & Left & Right & Stop \\
\midrule
Train & 15099 (53.41\%) & 2347 (8.30\%) & 2319 (8.20\%) & 8507 (30.09\%) \\
Test  & 4761  (58.38\%) & 698  (8.56\%) & 677  (8.30\%) & 2019 (24.76\%) \\
\bottomrule
\end{tabular*}
\end{table}

\subsection{Environment Construction and Robustness Setup}
\label{sec:env_setup}
Robustness under domain shift is evaluated using discrete environment labels.
Each sample is assigned an environment index $e\in\{1,\dots,E\}$ defined by the joint condition of precipitation intensity and part of day.
Precipitation intensity is discretised into three ordered levels using quantile binning on the training split, which yields balanced partitions while preserving the ordering from light to heavy precipitation.
Part of day is discretised into four six-hour intervals in local time: 00:00--06:00, 06:00--12:00, 12:00--18:00, and 18:00--24:00.
The nominal cross-product comprises up to twelve precipitation--time categories; the original experiment uses the environment indices produced by its data preparation pipeline.

The realized environment distribution is long tailed.
In the training split, the largest realized group contains 4{,}396 samples, while the smallest realized group contains 97 samples, corresponding to a ratio of approximately 45:1 between the largest and smallest groups.
In the test split, the largest realized group contains 1{,}946 samples, while the smallest realized group contains 74 samples, corresponding to a ratio of approximately 26:1.
Some environment codes appear in training but are absent in testing, and several realized groups remain rare with fewer than one hundred samples.

These environment labels are used in two ways.
First, for invariance learning, each training batch is partitioned into environment-specific subsets $\{\mathcal{B}_e\}_{e=1}^{E}$, and the IRM-style penalty in Eq.~\ref{eq:irm} is computed from the corresponding environment-wise risks $R_e$.
Second, for robustness evaluation, a leave-one-environment-out protocol is adopted: one large realised environment is held out as test data and the remaining environments are used for training.
This evaluates generalisation to an unseen group defined by precipitation and time-of-day metadata under a naturally imbalanced environment distribution; the setup does not isolate the effect of either environmental factor.

\subsection{Evaluation Protocols}
\label{sec:protocols}

\paragraph{IID evaluation}
IID evaluation uses the provided train split for training and the provided test split for evaluation.
In addition to accuracy, worst group accuracy is reported, defined as the minimum accuracy across realized environment groups.

\paragraph{Leave one environment out evaluation}
Leave one environment out evaluation holds out a target environment group as the test domain and trains on the remaining environment groups.
Due to the long tailed distribution, exhaustive leave one environment out across all groups can be unstable when the held out group is small.
A selected leave-one-environment-out protocol is therefore used: the selected held-out environment is described as the largest realised group, providing a more populated evaluation set while limiting the conclusions to this particular environment shift.

\paragraph{Two stage training}
A two stage pipeline is used.
Stage one is linear probing, where the encoder is frozen and a linear classifier is trained to probe representation quality.
Stage two is global finetuning, where the full model is finetuned end to end for final downstream performance.
Baseline comparisons are reported using stage-two fine-tuning, whereas component ablations use stage-one linear probing. The two protocols must not be interpreted as directly interchangeable.

\subsection{Baselines, Metrics, and Implementation Details}
\label{sec:baselines_metrics}

\paragraph{Baselines}
CARE is compared with DLinear~\cite{zeng2022transformerseffectivetimeseries}, LiST~\cite{Huang2024list}, PatchTST~\cite{nie2023timeseriesworth64}, TCN~\cite{bai2018empiricalevaluationgenericconvolutional}, DAE~\cite{al-falouji2024enhancing} and EAE~\cite{Gao2025FEAE}.
All baselines follow the same splits and evaluation protocols.

\paragraph{Metrics}
Accuracy and worst group accuracy measure predictive performance.
Negative log likelihood and Brier score evaluate probability quality.
Calibration is assessed using expected calibration error (ECE),
\[
\mathrm{ECE}=\sum_{m=1}^{M}\frac{|\mathcal{I}_m|}{N}\left|\mathrm{acc}(\mathcal{I}_m)-\mathrm{conf}(\mathcal{I}_m)\right|,
\]
where $\mathcal{I}_m$ denotes the index set of samples whose predicted confidence falls into the $m$-th bin, and $N$ is the number of evaluated samples.

\paragraph{Implementation details}
Unless otherwise stated, a Transformer backbone is used with hidden dimension $d_{\text{model}}=64$, feed-forward dimension $d_{\text{ff}}=128$, and number of encoder layers $e_{\text{layers}}=2$.
Batch size is 16 and the random seed is 2.
Self supervised pretraining uses 50 epochs with learning rate $10^{-4}$.
Stage one linear probing uses 5 epochs with learning rate $5\times10^{-3}$.
Stage two global finetuning uses 5 epochs with learning rate $10^{-4}$.
The past window length is $60$ and the total sequence length is $100$.

\subsection{Downstream Performance after Global Fine-Tuning}
\label{sec:exp_sota}

\paragraph{IID results}
Table~\ref{tab:sota_iid} reports stage two IID results.
CARE leads the reported IID fine-tuning comparison on accuracy (0.840), worst-group accuracy (0.800), ECE (0.023), NLL (0.453), and Brier score (0.233). These are downstream classification metrics, rather than direct measurements of the NIG head.

\begin{table}[t]
\centering
\caption{IID results using stage two global finetuning. Higher is better for accuracy and worst group accuracy. Lower is better for calibration and probability quality metric. Best values per column are in bold.}
\label{tab:sota_iid}
\small
\setlength{\tabcolsep}{4pt}
\renewcommand{\arraystretch}{1.11}
\begin{tabular*}{0.94\linewidth}{@{\extracolsep{\fill}}lccccc@{}}
\toprule
Model & Acc & Worst group & ECE & NLL & Brier \\
\midrule
CARE        & \textbf{0.840} & \textbf{0.800} & \textbf{0.023} & \textbf{0.453} & \textbf{0.233} \\
DLinear     & 0.821          & 0.794          & 0.044          & 0.539          & 0.254 \\
LiST        & 0.830          & 0.782          & 0.030          & 0.459          & 0.236 \\
PatchTST    & 0.817          & 0.774          & 0.070          & 0.544          & 0.269 \\
TCN         & 0.771          & 0.738          & 0.025          & 0.717          & 0.356 \\
DAE         & 0.791          & 0.770          & 0.041          & 0.677          & 0.334 \\
EAE         & 0.795          & 0.771          & 0.043          & 0.665          & 0.327 \\
\bottomrule
\end{tabular*}
\end{table}

\paragraph{Leave one environment out results}
Table~\ref{tab:sota_loo} reports stage two representative leave one environment out results.
CARE leads the reported selected held-out-environment comparison in accuracy (0.806) and Brier score (0.308).
In terms of negative log-likelihood, CARE is competitive and essentially tied with the strongest baseline (0.623 vs.\ 0.622).
TCN has the lowest ECE (0.039, compared with CARE's 0.090), so this evaluation supports a trade-off rather than a uniformly best-calibrated model.

\begin{table}[t]
\centering
\caption{Representative leave one environment out results using stage two global finetuning. The held out environment is chosen from the largest environment groups due to long tailed environment distribution. Higher is better for accuracy. Lower is better for calibration and probability quality metrics. Best values per column are in bold.}
\label{tab:sota_loo}
\small
\setlength{\tabcolsep}{4pt}
\renewcommand{\arraystretch}{1.11}
\begin{tabular*}{0.94\linewidth}{@{\extracolsep{\fill}}lcccc@{}}
\toprule
Model & Acc & ECE & NLL & Brier \\
\midrule
CARE        & \textbf{0.806} & 0.090          & 0.623          & \textbf{0.308} \\
DLinear     & 0.794          & 0.095          & 0.756          & 0.333 \\
LiST        & 0.785          & 0.070          & \textbf{0.622} & 0.312 \\
PatchTST    & 0.779          & 0.124          & 0.719          & 0.350 \\
TCN         & 0.741          & \textbf{0.039} & 0.773          & 0.396 \\
DAE         & 0.774          & 0.053          & 0.753          & 0.377 \\
EAE         & 0.771          & 0.061          & 0.747          & 0.375 \\
\bottomrule
\end{tabular*}
\end{table}

\subsection{Component Analysis with Frozen Representations}
\label{sec:exp_ablation}

This experiment evaluates the contribution of CARE components to encoder representation quality using stage one linear probing.
The encoder is frozen and only a linear classifier is trained.
Three components are ablated: reconstruction constraint REC, invariance regularization IRM, and evidential regularization EVI with weight 0.05.

\paragraph{IID ablation}
Table~\ref{tab:ablation_iid} reports IID linear-probe results.
Adding reconstruction (REC) provides the largest single-component improvement, increasing accuracy and yielding the best probability quality, with the lowest NLL and the lowest Brier score.
Enabling all components (All) achieves the best accuracy and worst-group accuracy, while its ECE is higher than the best combined settings (REC+IRM and REC+EVI), which attain the lowest ECE.
Overall, REC is the main driver for probability quality, whereas the full configuration offers the strongest accuracy under the linear-probe protocol.

\begin{table}[t]
\centering
\caption{IID component ablation with a frozen encoder (linear probing). Base denotes contrastive NCE; REC, IRM and EVI denote reconstruction, IRM-style regularisation and auxiliary evidence consistency. Best values per column are bold (Acc/worst-group Acc $\uparrow$, NLL/Brier/ECE $\downarrow$).}
\label{tab:ablation_iid}
\small
\setlength{\tabcolsep}{4pt}
\renewcommand{\arraystretch}{1.11}
\begin{tabular*}{0.94\linewidth}{@{\extracolsep{\fill}}lccccc@{}}
\toprule
Setting & Acc & Worst group & NLL & Brier & ECE \\
\midrule
Base     & 0.783          & 0.755          & 0.695          & 0.342          & 0.040 \\
REC      & 0.803          & 0.778          & \textbf{0.625} & \textbf{0.302} & 0.053 \\
IRM      & 0.784          & 0.755          & 0.694          & 0.341          & 0.038 \\
EVI      & 0.783          & 0.756          & 0.696          & 0.343          & 0.038 \\
REC+IRM  & 0.792          & 0.768          & 0.661          & 0.327          & \textbf{0.028} \\
REC+EVI  & 0.792          & 0.768          & 0.662          & 0.327          & \textbf{0.028} \\
IRM+EVI  & 0.781          & 0.754          & 0.700          & 0.344          & 0.041 \\
All      & \textbf{0.804} & \textbf{0.782} & \textbf{0.625} & 0.303          & 0.047 \\
\bottomrule
\end{tabular*}
\end{table}

\paragraph{Leave one environment out ablation}
Table~\ref{tab:ablation_loo} reports selected leave-one-environment-out linear-probe results.
On the reported held-out environment, the full configuration (All) has the lowest ECE (0.049) and NLL (0.714), and matches REC on Brier score (0.356).
REC has the highest accuracy in this linear-probing ablation (0.779), versus 0.777 for All; the combination therefore provides a calibration/likelihood trade-off rather than uniformly improving every metric.

\begin{table}[t]
\centering
\caption{Held-out-environment component ablation with a frozen encoder (linear probing). Base denotes contrastive NCE; REC, IRM and EVI denote reconstruction, IRM-style regularisation and auxiliary evidence consistency. Best values per column are bold (Acc $\uparrow$, NLL/Brier/ECE $\downarrow$).}
\label{tab:ablation_loo}
\small
\setlength{\tabcolsep}{4pt}
\renewcommand{\arraystretch}{1.11}
\begin{tabular*}{0.94\linewidth}{@{\extracolsep{\fill}}lcccc@{}}
\toprule
Setting & Acc & NLL & Brier & ECE \\
\midrule
Base     & 0.765          & 0.750          & 0.381          & 0.066 \\
REC      & \textbf{0.779} & 0.720          & \textbf{0.356} & 0.068 \\
IRM      & 0.765          & 0.746          & 0.380          & 0.073 \\
EVI      & 0.765          & 0.749          & 0.381          & 0.070 \\
REC+IRM  & 0.774          & 0.731          & 0.370          & 0.069 \\
REC+EVI  & 0.774          & 0.731          & 0.370          & 0.070 \\
IRM+EVI  & 0.762          & 0.752          & 0.382          & 0.075 \\
All      & 0.777          & \textbf{0.714} & \textbf{0.356} & \textbf{0.049} \\
\bottomrule
\end{tabular*}
\end{table}

\FloatBarrier
\subsection{Qualitative Inspection of Representation Geometry}
Beyond quantitative metrics, a qualitative inspection of the representation geometry learned by CARE is provided.
Embeddings are projected to two dimensions using UMAP~\cite{mcinnes2020umapuniformmanifoldapproximation}. Each point corresponds to one trajectory window of length $L{=}100$ with $F{=}7$ per-timestep features.
Points are coloured by attributes derived from the input window to assess whether the embedding manifold reflects behaviour-relevant structure.

\begin{figure}[!tbp]
  \centering
  \includegraphics[width=0.55\linewidth]{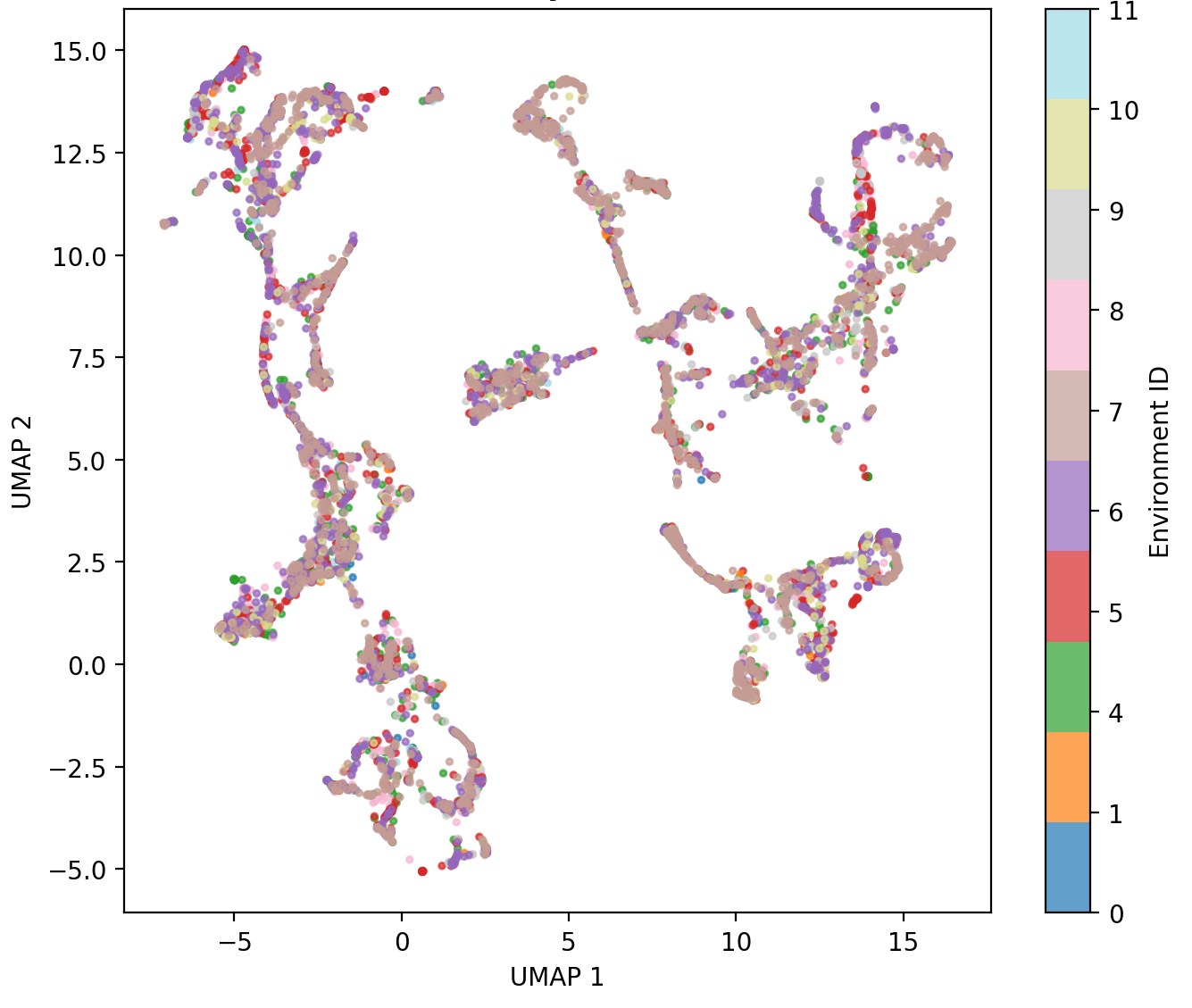}
  \caption{UMAP visualisation of CARE encoder embeddings on the test split, coloured by the discrete environment ID (used by CARE for environment-aware learning and evaluation).
  Colours appear interleaved in this two-dimensional projection; this alone does not establish environment invariance. Environment IDs 2 and 3 are absent in the test split and therefore do not appear in the plot.}
  \label{fig:care_umap_env}
\end{figure}

\begin{figure}[!tbp]
  \centering
  \includegraphics[width=0.98\linewidth]{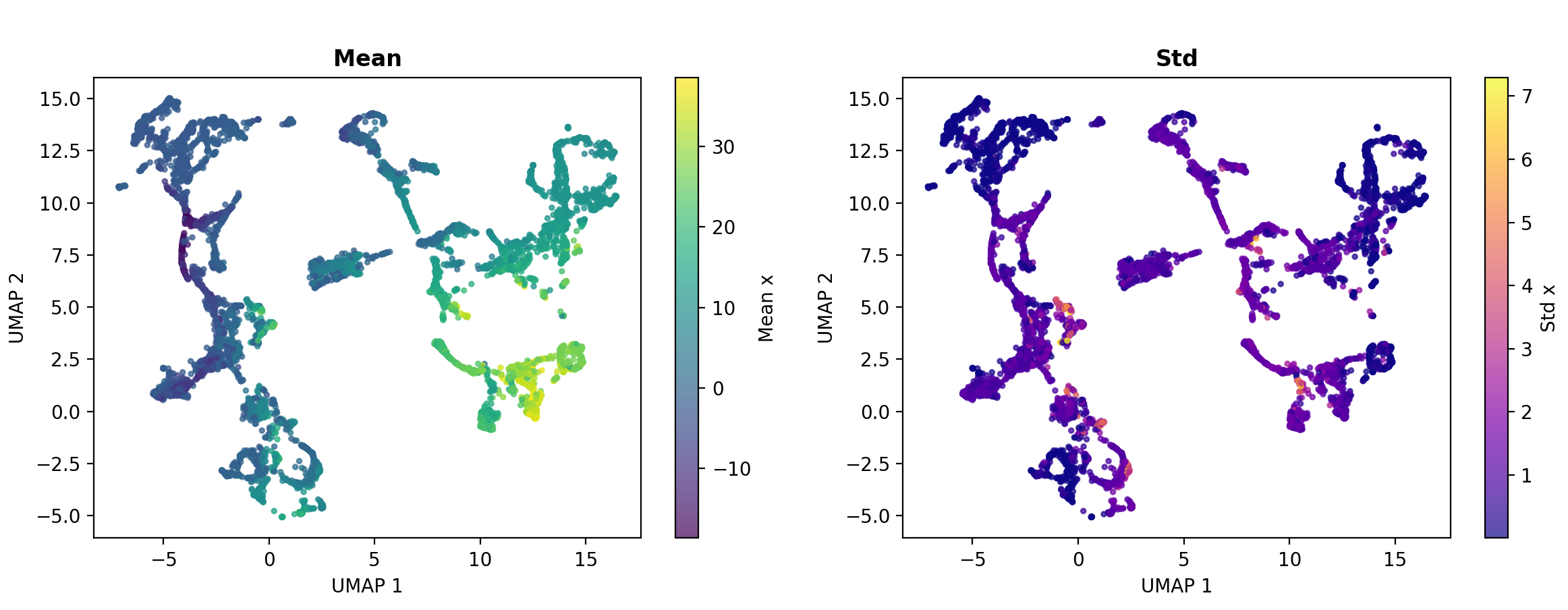}
  \caption{UMAP visualisation of CARE encoder embeddings on the test split, coloured by trajectory-level statistics of the $x$ coordinate over each window ($L{=}100$).
  Left: $\mathrm{mean}(x)$; right: $\mathrm{std}(x)$.
  The smooth gradient in $\mathrm{mean}(x)$ suggests coherent geometric organisation, while localised high $\mathrm{std}(x)$ highlights windows with larger spatial variability.}
  \label{fig:care_umap_x}
\end{figure}

\begin{figure}[!tbp]
  \centering
  \includegraphics[width=0.98\linewidth]{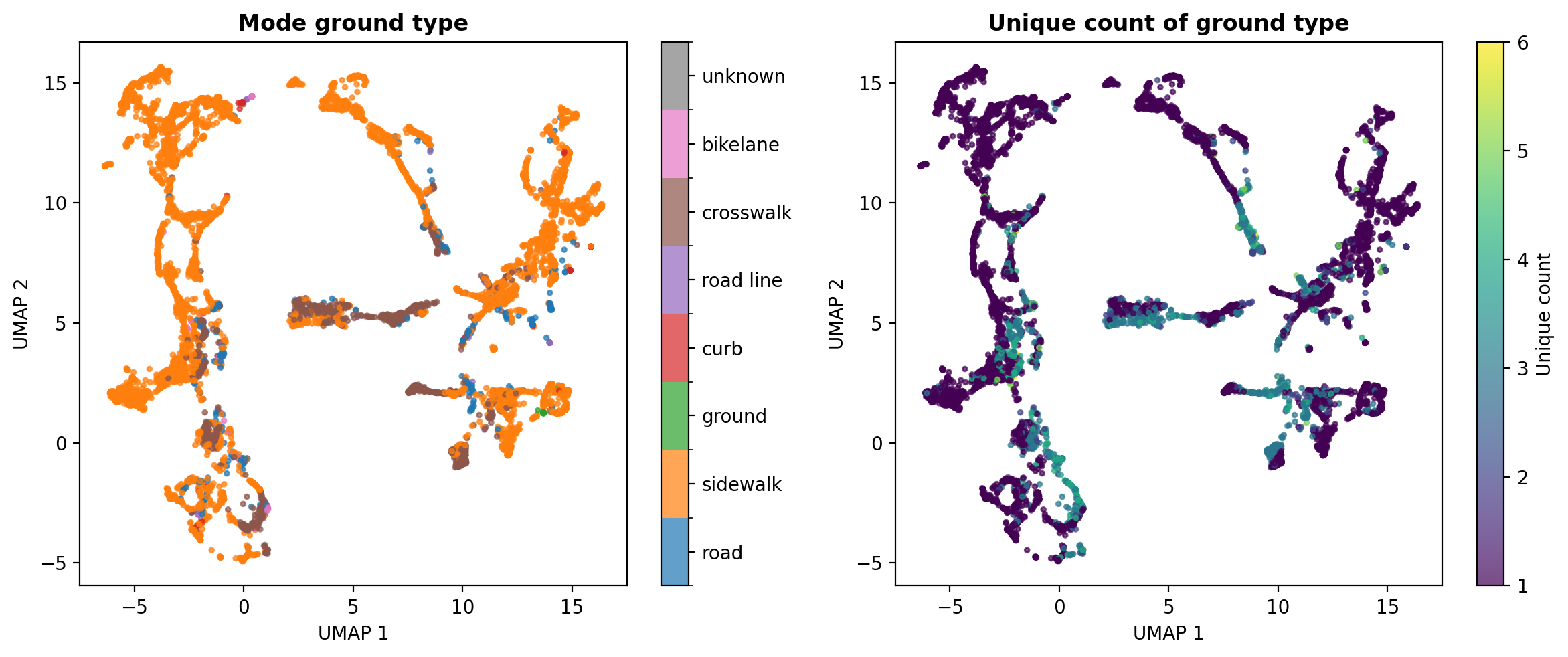}
  \caption{
    UMAP projection of CARE encoder embeddings on the test split, coloured by ground-type context aggregated within each trajectory window ($L{=}100$).
    Left: modal \texttt{ground\_type} in the window.
    Right: number of unique \texttt{ground\_type} values observed in the window, serving as a proxy for local context diversity.
    The dominant mode is consistent with typical VRU trajectories occurring on sidewalks or road surfaces.
    Colours remain largely interleaved across the manifold rather than forming isolated semantic islands, while regions with higher context diversity concentrate locally and align with areas of higher motion variability in other statistics, providing interpretable context cues without the representation being driven by discrete ground-type labels.
    }
    \label{fig:umap_ground}
\end{figure}

Fig.~\ref{fig:care_umap_env} shows that samples from different environments are largely interleaved across the embedding manifold, without forming clearly environment-separated islands.
This descriptive visualization does not show obvious environment-separated islands, but UMAP projections alone cannot establish statistical independence of features and environment labels.
Figure~\ref{fig:care_umap_x} colours the same embeddings by trajectory-level statistics of the $x$ coordinate.
A smooth gradient in $\mathrm{mean}(x)$ over the manifold suggests coherent geometric organisation, while localised regions with high $\mathrm{std}(x)$ highlight windows with larger motion variability.
Finally, Figure~\ref{fig:umap_ground} colours embeddings by ground-type context.
The modal \texttt{ground\_type} is dominated by the most common VRU surfaces and does not partition the manifold into isolated semantic clusters.
Instead, regions with higher ground-type diversity tend to concentrate in specific parts of the manifold and align with areas of higher motion variability, suggesting that CARE preserves interpretable scene-context cues in addition to continuous geometric factors.
These qualitative patterns suggest locally interpretable geometry. They do not constitute a quantitative test of environment invariance.

\section{Discussion}
\label{sec:discussion}
\subsection{Predictive Performance and Component Contributions}
The reported global fine-tuning comparison favours CARE on all five IID measures, including accuracy of 0.840 and ECE of 0.023. In the selected held-out-environment test, CARE achieves the highest reported accuracy (0.806) and the lowest Brier score (0.308), but TCN has a lower ECE (0.039 versus 0.090) and LiST has a marginally lower NLL (0.622 versus 0.623). These distinctions matter: a strong score under one particular environment shift does not imply invariance or uniformly superior calibration.

Component ablations tell a more selective story. With a frozen encoder, masked reconstruction gives the clearest individual gain in held-out-environment accuracy (0.779 versus 0.765 for the contrastive baseline). IRM alone matches the baseline in held-out-environment accuracy and has higher ECE, while EVI alone does not improve that accuracy. The full configuration improves held-out-environment ECE to 0.049 and NLL to 0.714 at accuracy 0.777, slightly below the 0.779 of REC. This pattern motivates studying interactions among regularizers; it does not justify treating all auxiliary penalties as independently beneficial.

\subsection{Limitations and Reproducibility}
The benchmark uses rule-derived proxy intention labels and the reported environment groups are long-tailed. The reported leave-one-environment-out results represent one fixed, large held-out group, not an exhaustive average across all environment groups. Experiments use the reported random seed of 2 and do not quantify seed-to-seed variability. UMAP visualizations provide qualitative interpretation but cannot establish learned invariance. The temporal perturbations affect the later portion of the \emph{input window}; the paper makes no causal intervention or future-label-leakage claim. The exact correspondence between numeric environment codes and the nominal precipitation--time categories is not fully documented in the available manuscript, limiting independent verification of environment group membership.

Future work could evaluate all adequately populated held-out environment groups, quantify multiple-seed uncertainty, and isolate the temporal-asymmetry and environment-risk penalty contributions under otherwise matched protocols.

\section{Conclusion}
CARE studies an environment-regularised contrastive objective for motion representation learning, combining temporally asymmetric views with an IRM-style penalty on environment-wise risks. Reconstruction and evidence consistency complement this main objective as auxiliary losses. On the IMPTC-derived intention benchmark, CARE obtains 0.840 IID accuracy and 0.806 accuracy on the selected held-out environment after fine-tuning. Under frozen-encoder evaluation, reconstruction is the strongest individual contributor to held-out accuracy, whereas the full objective provides the lowest ECE among the tested combinations. These results support the practical value of the proposed combination under the evaluated conditions, but do not establish causal invariance, broad environment-generalisation guarantees, or calibrated NIG uncertainty. The study provides an environment-wise contrastive regularisation formulation and a component-level evaluation of its predictive trade-offs in behavioural intention prediction.

\section*{Acknowledgments}
This research has been partly funded by the Hessian State Ministry for Higher Education, Research, Science and the Arts (HMWK) as part of the LOEWE program (Kapitel 15 02, F\"orderbuchungskreis 2995, F\"orderprodukt 11 -- Landes-Offensive zur Entwicklung Wissenschaftlich-\"okonomischer Exzellenz).

\bibliographystyle{unsrtnat}
\bibliography{refs}

\end{document}